%% file: samplepaper.tex
\documentclass[runningheads]{llncs}
\usepackage[T1]{fontenc}
\usepackage{graphicx}

\usepackage{tikz}
\usepackage{comment}
\usepackage{amsmath,amssymb} 
\usepackage{color}

\usepackage{sidecap}
\usepackage{booktabs}
\usepackage{mathtools}
\usepackage{dsfont}
\usepackage{multicol}
\usepackage{multirow}
\usepackage{subcaption}

\DeclareMathOperator*{\argmax}{argmax}

\begin{document}
%
\title{Continual Source-Free \\ Unsupervised Domain Adaptation}
%
%
\author{Waqar Ahmed\inst{1}\orcidID{0000-0001-5416-149X} \and
Pietro Morerio\inst{1}\orcidID{0000-0001-5259-1496} \and
Vittorio Murino\inst{1,2}\orcidID{0000-0002-8645-2328}}
\authorrunning{F. Author et al.}
%
\institute{Pattern Analysis \& Computer Vision, Istituto Italiano di Tecnologia, Genova, Italy 
\and Dipartimento di Informatica, University of Verona, Italy\\
\email{\{waqar.ahmed, pietro.morerio, vittorio.murino\}@iit.it}}
\maketitle              
\begin{abstract}
Existing Source-free Unsupervised Domain Adaptation (SUDA) approaches inherently exhibit catastrophic forgetting. Typically, models trained on a labeled source domain and adapted to unlabeled target data improve performance on the target while dropping performance on the source, which is not available during adaptation. In this study, our goal is to cope with the challenging problem of SUDA in a continual learning setting, \textit{i.e.,} adapting to the target(s) with varying distributional shifts while maintaining performance on the source. The proposed framework consists of two main stages: i) a SUDA model yielding cleaner target labels --- favoring good performance on target, and ii) a novel method for synthesizing class-conditioned source-style images by leveraging only the source model and pseudo-labeled target data as a prior. An extensive pool of experiments on major benchmarks, \textit{e.g.,} PACS, Visda-C, and DomainNet demonstrates that the proposed Continual SUDA (C-SUDA) framework enables preserving satisfactory performance on the source domain \emph{without} exploiting the source data at all.

\end{abstract}

\section{Introduction}
\label{sec:intro}
Convolutional Neural Networks (CNNs) trained on a labeled \textit{source} domain often fail to generalize well on a related but different \textit{target} domain due to the well-known \textit{domain shift} \cite{torralba2011unbiased,morerio2020generative}. Since annotating data from a new domain is expensive and sometimes even impossible, Unsupervised Domain Adaptation (UDA) methods have been developed to address the drop in performance by exploiting \textit{unlabelled} target data.

Conventional UDA methods address the adaptation task by \textit{e.g.,} feature alignment \cite{chen2019progressive}, matching moments \cite{peng2019moment}, 
or adversarial learning \cite{tang2020discriminative}. However, these methods typically require joint access to both \textit{labeled source} and \textit{unlabeled target} data during adaptation, making them unsuitable for most real-world scenarios, where source data is inaccessible (\textit{e.g.}, due to data privacy or proprietary reasons). Nevertheless, the performance on the source often degrades after adaptation to the target, even when using the source data during the training/adaptation process.

\begin{figure*}[t!]
\captionsetup{font=footnotesize}
\centering
\includegraphics[width=0.99\textwidth]{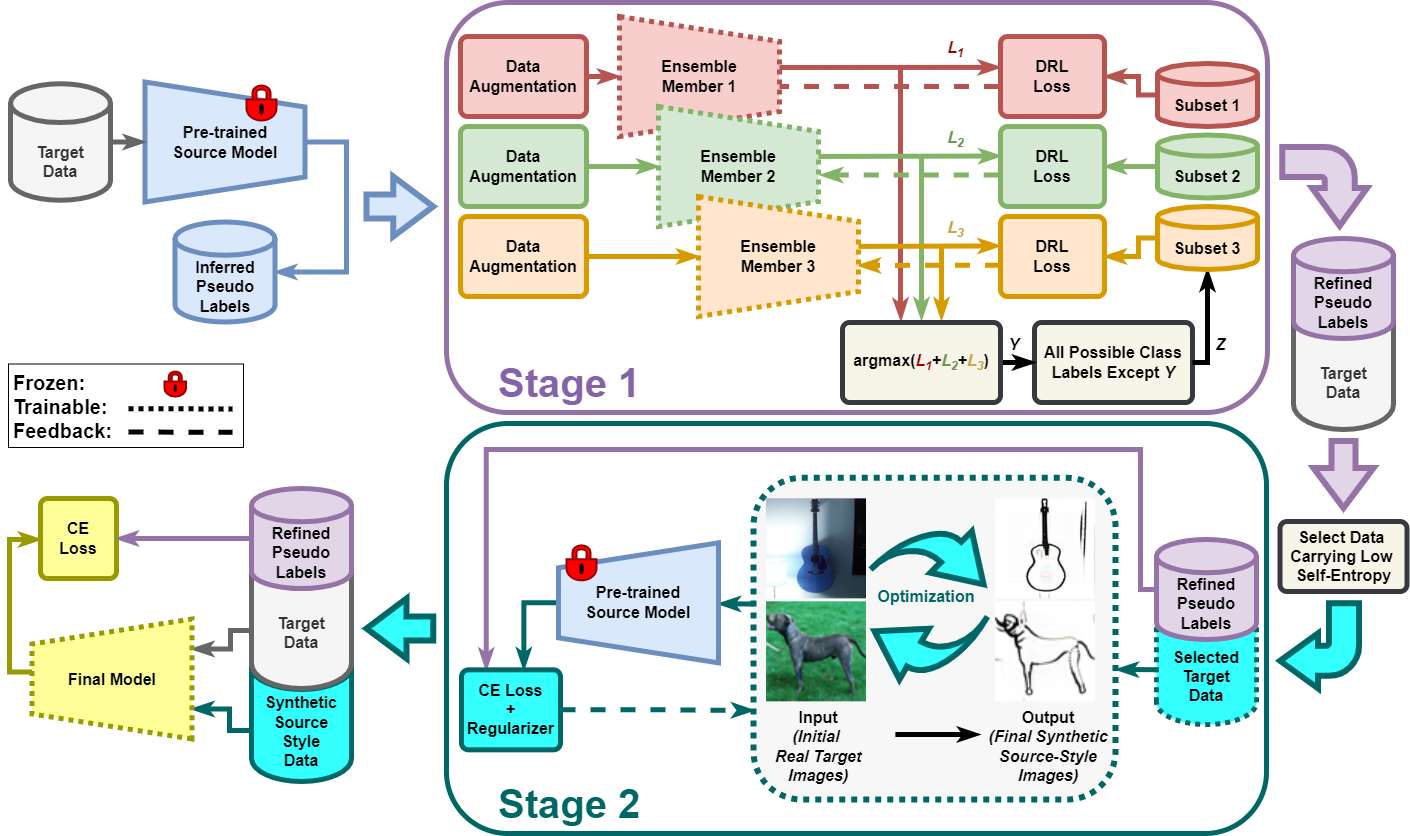}
\caption{\footnotesize Overview of the proposed continual Source-free Unsupervised Domain Adaptation (SUDA) method. We assume a pre-trained source model to infer pseudo-labels of the target set (top-left). \textit{Stage 1} refines incorrect pseudo-labels to achieve SUDA. \textit{Stage 2} synthesizes source-style images to avoid catastrophic forgetting of the source, thus achieving continual adaptation. Finally, a single model is trained using real target and synthetic source images, each one associated with a refined pseudo-label (bottom-left).}
\label{fig:HowWeDo}
\end{figure*}
 
Recently proposed methods address UDA problem under a more realistic source-free assumption, \textit{i.e.}, by using the pre-trained source model and unlabeled target domain data only \cite{li2020model,morerio2020generative,pmlr-v119-liang20a}. However, due to absolutely no exposure to source data distribution by any means, these methods naturally undergo catastrophic forgetting, \textit{i.e.}, the model adapted to the target experiences a substantial drop in performance if tested back on the source domain data.

This is actually a severe drawback in a practical realistic scenario. For example, a self-driving car company releases a recognition model trained on some proprietary data, \textit{e.g.}, data collected in unknown urban regions, considered as a source. 
Thereafter, one will have to perform Source-free Unsupervised Domain Adaptation (SUDA) to obtain acceptable performance on a different target domain \textit{e.g.,} rural or mountain regions. Yet, due to the unavoidable catastrophic forgetting issue, the adapted model will likely fail when deployed back in urban areas. The availability of multiple models trained on different domains would not work since it is clearly neither feasible nor scalable due to the typical limited hardware resources. Thus, it is desirable to preserve the model's performance on the source domain too for practical use cases.


To address this challenge, we propose a Continual Source-free Unsupervised Domain Adaptation (C-SUDA) framework to preserve satisfactory performance on the source while adapting to the target domain, yet, assuming no access to source data at all. The C-SUDA is composed of two main stages (See Fig.~\ref{fig:HowWeDo}).

\textit{Stage 1} assumes a SUDA model that can provide (ideally) correct \textit{pseudo-labels} of the unlabeled target samples.  
One possibility is to infer the required pseudo-labels using a model pre-trained on the source domain (see top-left of Fig. \ref{fig:HowWeDo}). However, this results in a significant amount of incorrect/noisy pseudo-labels --- a consequence of the \textit{domain shift} \cite{torralba2011unbiased,morerio2020generative} --- which need to be refined. 
In this study, we leverage a related SUDA method proposed in a previous work \cite{ahmed2022cleaning} which refines the inferred pseudo-labels considering single-source 
, multi-source 
, or multi-target scenarios indifferently.
\cite{ahmed2022cleaning} is an \textit{ensemble learning} method aimed at training ensemble members with randomly sampled \textit{disjoint} subsets of residual labels. This allows each member to learn something diverse and possibly complementary, leading to a stronger consensus and noise resilience. Consequently, it progressively separates samples carrying noisy pseudo-labels from the clean ones, by assigning them lower and higher class confidence, respectively. Eventually, the low-confidence samples iteratively undergo a pseudo-label refinement process via reassignment. Note however that any of-the-shelf SUDA methods can be used here which provides clean pseudo-labels.

\textit{Stage 2} consists of an image synthesis process aimed at promoting continual learning, by generating synthetic source images that help in preserving satisfactory performance on the source domain. In particular, we leverage the fact that CNNs are capable of automatically discovering the rich underlying patterns hidden in the data (\textit{e.g.}, the running average statistics stored in the Batch-Norm layers). The idea here is to employ feature distribution regularizers exploiting such rich information to generate class-conditioned (based on refined pseudo-labels obtained in Stage 1) source-style images. Only images with high confidence (low soft prediction entropy) are used in this stage.
Specifically, we use target images as prior and optimize their statistics and style such that the transformed versions of the generated images resemble the source domain distribution and achieve sharp classification predictions when presented to the source model.
We mix synthetic-source and real-target images for training the final (single) model that ensures good performance on both \textit{real} source and target domains.

The proposed C-SUDA framework is fully-adaptive and can cope with single-source, multi-source, and multi-target UDA problems indifferently.
With extensive experiments on various benchmarks carrying different amounts of inferred pseudo-labels noise, we show that our framework helps alleviating catastrophic forgetting when the model is tested back on the source domain(s). 
To summarise, the contributions of our work can be stated as follows:

\begin{itemize} 
\setlength\itemsep{0pt}
    \item We propose a new, versatile Continual Source-free UDA (C-SUDA)framework, composed of two simple, yet effective stages, which attains state-of-the-art performance in both target and source domains. Notably, our method can equally face single-source, multi-source, and multi-target UDA scenarios indifferently.
    \item The second stage is a new image synthesis method that leverages pseudo-labeled target images and a pre-trained source model to generate high-fidelity class-conditional source-style images. 
    Such synthetic images help preserving good performance on source domain(s) without using the real source samples.
    \item We validate our method on three well-known benchmarks, demonstrating the proposed C-SUDA's effectiveness and generalizability in diverse scenarios.
\end{itemize}

The remainder of the paper is organized as follows. Sec.~\ref{RW} discusses related works. Sec.~\ref{Method} introduces the proposed method. Sec.~\ref{EXP} presents the experimental setup and obtained results. 
Finally, conclusions are drawn in Sec.~\ref{Conclusion}.

\section{Related Work}
\label{RW}
Our work lies at the intersection of \textit{Source-Free UDA} and \textit{Continual Learning} (CL), whose related literature is discussed below. Also, we briefly review existing literature related to 
\textit{Image Synthesis}, as synthesized source images constitute a core stage of the proposed pipeline.

\paragraph{Source-Free UDA.}
Recent years have seen growing interest in addressing the UDA problem in realistic source-free settings. A common approach is to leverage a pre-trained source model for either transferring the fixed source classifier to the target data employing information maximization and pseudo-labeling \cite{pmlr-v119-liang20a},
or updating target model progressively by generating target-style samples through conditional generative adversarial networks \cite{morerio2020generative}, also by combining clustering-based regularization \cite{li2020model}. Similarly, to improve performance in domain-adaptive person re-identification tasks, \cite{ge2020mutual} proposes a pseudo-label cleaning process with online refined soft pseudo-labels. Our proposed approach lies in this category of works, but 
we differ from previous methods by not requiring: i) a customized network, ii) to generate target-style data using \textit{e.g.}, GAN-based models that require careful hyper-parameter tuning to reach stability, and iii) different techniques to tackle single-source, multi-source, and multi-target UDA. 

\paragraph{Continual Learning.}
CL refers to 
learning by using many diverse data distributions (tasks) sequentially, while avoiding catastrophic forgetting \cite{bang2021rainbow,shi2021continual}.
Some recent works tackled the continual UDA problem such as, \textit{e.g.,} 
\cite{Bobu2018AdaptingTC}, which proposes domain adversarial learning with sample replay. More recently, 
\cite{Volpi_2021_CVPR} employs a meta-learning strategy and domain randomization using heavy image manipulations. Despite the fact that these strategies are effective, they all require the access to source data. 
Similarly, \cite{yang2021generalized} performs continual adaptation, but needs two auxiliary binary embedding layers to be specifically trained on source. 
These are stored and used during adaptation in order to overcome catastrophic forgetting, and this questions the source-free nature claimed by the method. On the contrary, we only assume the availability of standard CNN models pre-trained on the source, \textcolor{black}{but do not require retraining of any auxiliary layers on source nor the access to partial source data to tackle the catastrophic forgetting.}

\paragraph{Image Synthesis.}
In the deep learning era, this area refers to the generation of synthetic images, possibly indistinguishable from real ones, and generative adversarial networks (GAN) are among the most popular class of approaches adopted to date.
GAN-inversion is an interesting research direction, in which an anchor image is used to guide a GAN to generate realistic images by inverting a pre-trained model \cite{pan2021exploiting}.
Other works focus on network inversion that enables noise-to-image transformation by back-propagating gradients to the learnable input images like in, \textit{e.g.}, \cite{mordvintsev2015inceptionism}, which introduces “dreaming” new visual features onto images, while \cite{NEURIPS2019_6f2268bd} takes this approach a step further to generate more realistic images.
In the context of Domain Generalization, \cite{volpi2018generalizing} optimizes images in the pixels space to produce augmentations, extending the working domain of the classifier to unseen data distributions. 
\textcolor{black}{
Recently proposed Deep-Inversion \cite{yin2020dreaming} method assumes a uniform prior and backpropagates desired label to produce synthetic images using a regularizer based on feature matching for 
data-free knowledge distillation. 
We took inspiration from such work, but we realized that such synthetic images were not helpful in preserving source performance. Differently, we cast the problem as a conditional image-to-image translation task. The notable impact is that, with the refined pseudo-labels obtained in Stage 1, target samples can be converted to synthetic source-style images that are not only visually plausible, but also suitable for training a classifier. 
}


\section{The C-SUDA Method}
\label{Method}
Our method comprises two main stages (see Fig. \ref{fig:HowWeDo}). After inferring pseudo-labels of the target samples --- using the source model, Stage 1 is devoted to \textit{refining} the so called shift-noise \cite{morerio2020generative} (affecting target samples) resulting in cleaner pseudo-labels. The strategy proposed in \cite{ahmed2022cleaning}, detailed in Section \ref{subsec:source-f}, is sufficient to obtain state-of-the-art performance on the target set in the source-free setting, yet nothing prevents catastrophic forgetting. For this reason, the target set with refined pseudo-labels is exploited in Stage 2 as a prior for synthesizing source-style images which can ``anchor'' the model to its original performance on the source set, as detailed in Section \ref{subsec:image-syn}.

\paragraph{Preliminaries.}
The goal of UDA is to adapt a model pre-trained on a labelled source domain $\mathcal{D}_s=\{(\boldsymbol{x}^i_s,y^i_s)\}_{i=1}^{N_s}$ 
on a different, yet related, unlabeled target domain $\mathcal{D}_t=\{\boldsymbol{x}^j_t\}_{j=1}^{N_t}$, which share the same label set, \textit{i.e.,} $\mathcal{Y}_s = \mathcal{Y}_t$.
We assume $\mathcal{D}_s$ is \textit{never} available in a realistic source-free scenario, while of course a pre-trained source model $f_s(\cdot)$ is at our disposal. This can be used to infer pseudo-labels $\mathcal{P}={\{\tilde{y}^j \}}_{j=1}^{N_t}$ for the target domain: 
\begin{equation}
    \tilde{y}^j = \argmax f_s(\boldsymbol{x}^j_t), \; j=1...N_t
\end{equation}
Clearly, $\mathcal{P}$ would be noisy, i.e., a significant amount of pseudo-labels is wrong due to domain shift. The following section discusses how to progressively filter out such noise and subsequently obtains a cleaner set $\mathcal{P'}$, which in turn translates into better accuracy on the target set.

\subsection{Stage 1: Pseudo-label refinement}\label{subsec:source-f}
Training a target model with cross-entropy given $\mathcal{P}$ as supervisory signal (i.e., using $\mathcal{P}$ as ground-truth) eventually results in overfitting noisy samples. For instance, in case of wrong pseudo-labels $\tilde{y}^j \neq y_t^j$ ($y_t^j$ are \textit{unknown} target labels), the model would undeniably try to maximize the probability of a sample belonging to the wrong class.

To mitigate such a problem, an Ensemble of classifiers can be used (see Fig. \ref{fig:HowWeDo}), which denotes the class of techniques of concurrently training multiple networks and, in its simplest form, averaging their output. 
\textcolor{black}{More in detail, it adopts an idea of employing stochastically sampled \textit{Disjoint Residual Labels (DRL)}: equally distributed disjoint subsets of complementary labels (spanning entire class-set except the given pseudo-label $\tilde{y}^j$) are used to back-propagate different feedback to each ensemble member. 
This helps ensemble members to learn different concepts and to achieve a strong consensus.
For instance, in case of wrong pseudo-label $\tilde{y}^j$, the correct label $y_t^j$ is wrongly provided as one of the complementary labels to only one member, while other members always learn from clean feedback. Thus, the loss is defined as:} 
\begin{equation}\label{eq:nl-RL}
\mathcal{L}_{\small{DRL}}(\mathcal{D}_t)=-\mathbb{E}_{x_t \sim\mathcal{D}_t} \frac{1}{N_{DRL}}\sum_{c=1}^{C} \mathds{1}_{[c \in DRL]}log(1-p^c)
\end{equation}
which is used to train each member \textit{independently}. The predictions of the members are then late-fused via:

\begin{equation}\label{eq:ens}
\boldsymbol{p}_e= \boldsymbol{\sigma}(\dfrac{1}{N_e\cdot N_a}\sum_{k=1}^{N_e} \sum_{l=1}^{N_a}
f^{k,a} (\boldsymbol{x}))
\end{equation}
where $f^k$ is one of the $N_e$ members and we use a moving average of $N_a$ previous outputs ($N_a=10$ for all the experiments).

With the growing number of epochs, noisy samples remain towards low confidence regime and clean samples obtain high confidence progressively. 
Consequently, pseudo-label refinement is achieved progressively during training in an adaptive manner,
where the total noise is progressively reduced). 



\subsection{Stage 2: Image Synthesis for Continual Adaptation}
\label{subsec:image-syn}
As a result of Stage 1, a refined pseudo-label is associated to each target sample. Note that, at this point, some noise still remains in pseudo-labels, \textit{i.e.}, some of them still do not correspond to the correct 
target label. 
Since we need to generate class-conditioned source-style images, by leveraging the source model together with $(\boldsymbol{x}^j_t, \tilde{y}^j_t)$ as a prior, we feed Stage 2 with only a subset of the target set which most likely guarantees less noise (see Fig. \ref{fig:HowWeDo}). More in detail, we compute prediction uncertainty of the target samples quantified by self-entropy \cite{9528982} as:
\begin{equation}\label{eq:self-ent}
    H(\boldsymbol{x}_t) = -\sum \boldsymbol{p}(\boldsymbol{x}_t)log(\boldsymbol{p}(\boldsymbol{x}_t)),
\end{equation}
where smaller entropy indicates more confident prediction (more details in supp. mat.). Based on this, we sort the target samples in ascending order and select only the first $N_h$ samples from each class.
These selected target samples are then optimized by minimizing the cross-entropy loss for the original source model and two feature distribution regularization terms. 
Note that here the source model (the only available information we have about the source domain) is kept frozen while we optimize pseudo-labeled target samples $(\boldsymbol{x}^j_t, \tilde{y}^j_t)$ in the pixel space:
\begin{equation}
    \boldsymbol{x} \leftarrow \boldsymbol{x} - \eta \nabla_{\boldsymbol{x}} \mathcal{L}(f_s(\boldsymbol{x}), \tilde{y}), \quad \boldsymbol{x}_0 = \boldsymbol{x}^j_t
\end{equation}
\begin{equation}
\begin{split}
    \mathcal{L}(f_s(\boldsymbol{x}), \tilde{y}) = & \ell_{CE}((f_s(\boldsymbol{x})), \tilde{y}) + \\ 
    & \lambda_{TV}\mathcal{R}_{TV}(\boldsymbol{x}) + \lambda_{BN}\mathcal{R}_{BN}(\boldsymbol{x}),
\end{split}
\end{equation}
\begin{equation*}\label{eq:TV}
\mathcal{R}_{TV}(\boldsymbol{x}) = \sum_{u,v}((\boldsymbol{x}_{u,v+1} - \boldsymbol{x}_{uv})^2 + (\boldsymbol{x}_{u+1,v} - \boldsymbol{x}_{uv})^2)^{\frac{1}{2}},
\end{equation*}
\begin{equation*}\label{eq:BN} 
    \mathcal{R}_{BN} = \sum_{l,j} \parallel \mu_l(\boldsymbol{x}^j) - \mu_{l}\parallel + 
    \parallel \sigma^2_l(\boldsymbol{x}^j) - \sigma_l^2) \parallel_2,
\end{equation*}
Here $\ell_{CE}(\cdot)$ is the cross-entropy loss, $f_s$ is the frozen source model, $\lambda_{TV}$ and $\lambda_{BN}$ are scalar weights. 

$\mathcal{R}_{TV}$ is a regularizer that penalizes the \textit{Total Variation norm} approximated as finite pixel difference ($u,v$ are pixel indexes)\cite{Mahendran_2015_CVPR}: it provides more stable convergence and encourages $\boldsymbol{x}$ to consist of piece-wise uniform patches.

$\mathcal{R}_{BN}$ enforces batch-wise ($j$ is the batch index) feature statistics similarities at all layers $l$, exploiting the BatchNorm running average parameters $(\mu_{l}, \sigma_{l})$ stored in the source models, which implicitly capture the channel-wise means and variances of the original source images \cite{yin2020dreaming}.

\begin{figure}[!t]
\centering
\includegraphics[width=\textwidth]{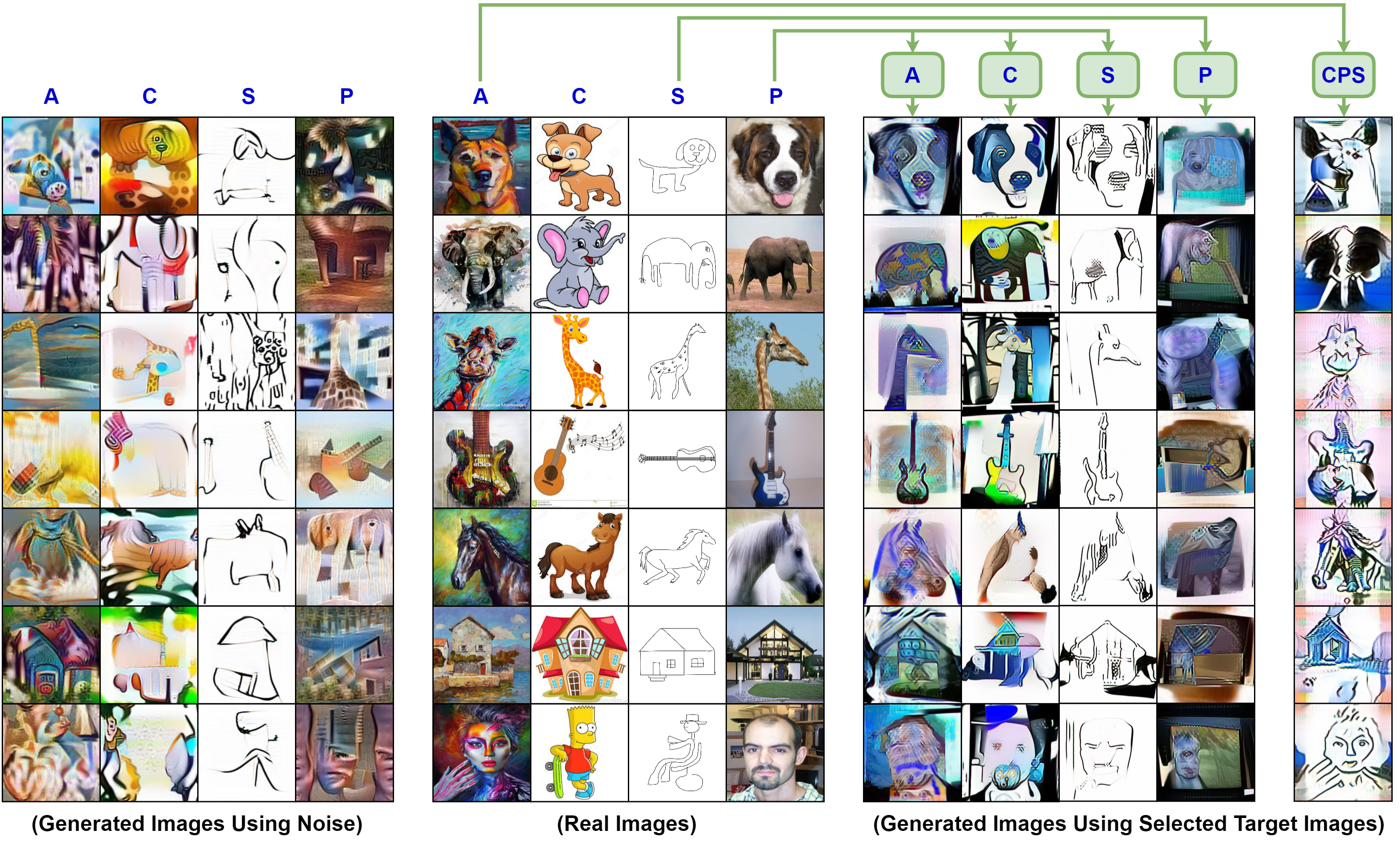}
\caption{\footnotesize  Image Synthesis using PACS dataset. Source style images (\textit{right}) are optimized from target images with refined pseudo-labels (\textit{center}). we also provide an example of images synthesized from random noise (\textit{left}). CPS exemplifies a multi-source case.
Legend:  \textit{\textcolor{blue}{\textbf{A}}: Art-painting, \textcolor{blue}{\textbf{C}}: Cartoon, \textcolor{blue}{\textbf{P}}: Photo, and \textcolor{blue}{\textbf{S}}: Sketch.}
}
\label{fig:images}
\end{figure}
In principle, images can be generated by optimizing random noise \cite{yin2020dreaming}. But we verified that starting from the actual target images guarantees higher fidelity, realism, and most of all diversity in generated synthetic images (see Fig.~\ref{fig:images}).
We also found that just a handful of synthetic source-style samples generated by our proposed method effectively helps in preserving source performance.

\noindent
\textbf{Final Model.}
\textcolor{black}{Once target data have been assigned pseudo-labels and source-style synthetic images are generated, we can proceed with the final model training (see Fig. \ref{fig:HowWeDo}, bottom left).}  We initialize the weights of the final trainable model with the source model. Subsequently, the weights of fully-connected (FC) layers are frozen while feature-extractor (FE) remains trainable. We train the final model with standard cross-entropy loss using both real target and synthetic source-style images together with corresponding refined pseudo-labels. Freezing FC layers not only help FE to learn the representation of synthetic source images most likely to be identical to the real source domain but also forces the target domain to get aligned with the source. 



%


\section{Experiments}
\label{EXP}
For the image classification task, we evaluate performance of our method via extensive experiments on major UDA benchmarks including, 
PACS \cite{li2017deeper}, VisDA-C \cite{peng2018visda}, and DomainNet \cite{peng2019moment}. 


%
For stage 1, we evaluate the refined pseudo-labels from \cite{ahmed2022cleaning}, which is based on the ensemble network comprising $3$ members. 
We examine the effectiveness of stage 2 i.e., the usefulness of synthesized images in preserving good performance on the real source domain. Note that in stage 2, we have access to the pre-trained source model and the target images with associated refined pseudo-labels (output of stage 1) only. For single-source and multi-source continual source-free UDA, we synthesis a relevant source for each target. For multi-target case, one synthesized source is enough for all related targets.

In all experiments related to image synthesis, we synthesize $32$ images per class---all together as one batch. The batch is initialized with real target samples carrying lowest self-entropy according to Eq.~\ref{eq:self-ent}. We use $Adam$ with a learning rate of $1e-1$ (with cosine annealing schedule) for images optimization. We set $\lambda_{TV}=1e-4$, $\lambda_{BN}=1e-2$, and batch receives $10K$ updates. Sample images are provided in Fig. \ref{fig:images}. 

\input{tables/table_PACS-SSDA_CSF-UDA}
\input{tables/table_PACS}

\subsection{Results}
The reported results in Tab.~\ref{tab:PACS_ssda}-\ref{tab:PACS} present average accuracy of 3 runs. 
In Table~\ref{tab:PACS_ssda}, we report results for all the possible pairs. Also, we report upper-bound performance (Baseline), however, we skip this information in the rest of the tables. 
In Tab.~\ref{tab:domainnet}-\ref{tab:PACS}, the \textit{Sc$\rightarrow$Tg} row reports the amount of correct target pseudo-labels acquired using the frozen source model. 
Along with the performance of SUDA and C-SUDA on target, we also report the performance \textit{degradation} on source due to source-free UDA (SUDA$^\star$) as well as the effectiveness of our method (C-SUDA$^\star$) in preserving the performance on the source.

\input{tables/table_domainnet}
\input{tables/table_VisdaC}

In Table~\ref{tab:PACS} \textit{(left)}, we compare our method with the existing approaches addressing  multi-target UDA on PACS. As can be noticed, with comparable performance in 2 cases, our method achieves superior average accuracy. For multi-source UDA, we compare recent works in Table~\ref{tab:PACS} \textit{(right)}. Also in this framework, our method consistently outperforms existing methods, with only in one case getting lower, yet comparable, accuracy. 

In Table~\ref{tab:domainnet}, with comparable performance in one case, C-SUDA consistently outperforms existing methods despite the large number of classes and discrepancy across domains.
Also in Table~\ref{tab:VisdaC}, the proposed method achieves state-of-the-art average accuracy on such a challenging benchmark. 
\section{Conclusions}
\label{Conclusion}
This work proposes Continual Source-Free Unsupervised Domain Adaptation as a realistic adaptation scenario where source samples are not available, but the performance is to be preserved on the source domain. 
Our method is composed of two stages, a Source-free UDA technique based on pseudo-label refinement, and a procedure for synthesizing source-style images to avoid catastrophic forgetting.
The proposed pipeline effectively solves the task by only assuming a pre-trained source model.
We empirically demonstrate that our proposed method achieves state-of-the-art performance on major UDA benchmarks.

\clearpage
%
%
\bibliographystyle{splncs04}
\bibliography{egbib}
\end{document}

%% file: tables/table_PACS-SSDA_CSF-UDA.tex
\setlength{\tabcolsep}{12.3pt}
\begin{table}[!t]
\centering
\scriptsize
\begin{tabular}{|c|c|c|c c | c c | c c|}
\hline
\multicolumn{2}{|c|}{Train} & 
Sc &
\multicolumn{2}{|c|}{Tg} &
\multicolumn{2}{|c|}{Tg+\textcolor{red}{\textbf{\textit{SynSc}}}} &
\multicolumn{2}{|c|}{Tg+Sc} 
\\
\hline
\multicolumn{2}{|c|}{Test} & 
Tg &
Tg &
\textit{Sc} &
Tg &
\textit{Sc} &
Tg &
\textit{Sc} 
\\
\hline
Sc & Tg & & \multicolumn{2}{|c|}{SUDA} & \multicolumn{2}{|c|}{C-SUDA} &  \multicolumn{2}{|c|}{Baseline}  \\
\hline
\multirow{3}{*}{\textcolor{blue}{\textbf{A}}} 
& \textcolor{blue}{\textbf{C}} & 58.1 & 84.3 & \textit{63.0} & 84.5 & \textit{81.0}
& 98.6 & \textit{98.1} \\
& \textcolor{blue}{\textbf{P}} & 96.0 & 98.4 & \textit{62.5} & 98.0 & \textit{75.9}
& 99.5 & \textit{98.7} \\
& \textcolor{blue}{\textbf{S}} & 43.9 & 56.2 & \textit{17.9} & 55.7 & \textit{65.5}
& 96.4 & \textit{98.6} \\
\hline
\multirow{3}{*}{\textcolor{blue}{\textbf{C}}}
& \textcolor{blue}{\textbf{A}} & 67.3 & 89.0 & \textit{61.9} & 88.5 & \textit{78.5}
& 98.0 & \textit{98.2} \\
& \textcolor{blue}{\textbf{P}} & 85.6 & 97.2 & \textit{21.2} & 96.4 & \textit{70.4}
& 98.9 & \textit{98.9} \\
& \textcolor{blue}{\textbf{S}} & 60.6 & 77.6 & \textit{46.9} & 77.3 & \textit{71.2}
& 96.6 & \textit{98.8} \\
\hline
\multirow{3}{*}{\textcolor{blue}{\textbf{P}}} 
& \textcolor{blue}{\textbf{A}} & 60.9 & 82.6 & \textit{87.9} & 83.1 & \textit{92.0}
& 98.1 & \textit{99.1} \\
& \textcolor{blue}{\textbf{C}} & 24.8 & 80.5 & \textit{68.2} & 80.6 & \textit{90.8}
& 99.1 & \textit{99.5} \\
& \textcolor{blue}{\textbf{S}} & 26.5 & 32.3 & \textit{13.4} & 33.2 & \textit{90.9}
& 96.2 & \textit{98.2} \\
\hline
\multirow{3}{*}{\textcolor{blue}{\textbf{S}}} 
& \textcolor{blue}{\textbf{A}} & 18.1 & 67.6 & \textit{42.6} & 67.2 & \textit{74.7}
& 97.8 & \textit{96.7} \\
& \textcolor{blue}{\textbf{C}} & 32.6 & 83.8 & \textit{52.8} & 83.9 & \textit{72.3}
& 99.0 & \textit{95.1} \\
& \textcolor{blue}{\textbf{P}} & 24.3 & 77.1 & \textit{17.9} & 77.0 & \textit{72.6}
& 99.6 & \textit{96.7} \\
\hline
\multicolumn{2}{|c|}{Avg.} & 49.9 & 77.2 & \textit{46.3} & 77.1 & \textit{78.0}
& 98.1 & \textit{98.0} \\
\hline
\end{tabular}
\caption{Classification accuracy on PACS with ResNet18. 
Legends: \textit{\textbf{Sc:} Source (real), \textbf{Tg:} Target (real), \textcolor{red}{\textbf{SynSc}}: Synthetic source (generated), \textbf{SUDA:} Source-free UDA, \textbf{C-SUDA:} Continual source-free UDA, 
\textcolor{blue}{\textbf{A}}: Art-painting, \textcolor{blue}{\textbf{C}}: Cartoon, \textcolor{blue}{\textbf{P}}: Photo, and \textcolor{blue}{\textbf{S}}: Sketch.}}
\vspace{-1.5em}
\label{tab:PACS_ssda}
\end{table}

%% file: tables/table_PACS.tex
\setlength{\tabcolsep}{1.4pt}
\renewcommand{\arraystretch}{1.2} 
\begin{table*}[ht]
\centering
\scriptsize
\begin{tabular}{|l|c c c c c c |c||l|c c c c |c|}
\hline
\multicolumn{8}{|c||}{\textbf{Multi-Target C-SUDA}} & \multicolumn{6}{c|}{\textbf{Multi-Source C-SUDA}} \\
\hline
Sc & 
\textcolor{blue}{\textbf{\textit{ }}} & 
\textcolor{blue}{\textbf{\textit{P}}} & 
\multicolumn{1}{ c| }{\textcolor{blue}{\textbf{\textit{ }}}} & 
\textcolor{blue}{\textbf{\textit{ }}} &
\textcolor{blue}{\textbf{\textit{A}}} &
\multicolumn{1}{ c| }{\textcolor{blue}{\textbf{\textit{ }}}} & 
\multirow{2}{*}{Avg.} & Sc
&
\textcolor{blue}{\textbf{\textit{C,P,S}}} & 
\textcolor{blue}{\textbf{\textit{A,P,S}}} &
\textcolor{blue}{\textbf{\textit{A,C,S}}} &
\multicolumn{1}{ c| }{\textcolor{blue}{\textcolor{blue}{\textbf{\textit{A,C,P}}}}} & 
\multirow{2}{*}{Avg.} 
\\
\cline{1-7}
\cline{9-13}
Tg & 
\textcolor{blue}{\textbf{\textit{A}}} &
\textcolor{blue}{\textbf{\textit{C}}} &
\multicolumn{1}{ c| }{\textcolor{blue}{\textbf{\textit{S}}}} & 
\textcolor{blue}{\textbf{\textit{P}}} &
\textcolor{blue}{\textbf{\textit{C}}} &
\multicolumn{1}{ c| }{\textcolor{blue}{\textcolor{blue}{\textbf{\textit{S}}}}} &
&
Tg
&
\textcolor{blue}{\textbf{\textit{A}}} &
\textcolor{blue}{\textbf{\textit{C}}} &
\textcolor{blue}{\textbf{\textit{P}}} &
\multicolumn{1}{ c| }{\textcolor{blue}{\textcolor{blue}{\textbf{\textit{S}}}}} & 
\\
\hline
ADDA* & 24.3 & 20.1 & 22.4 & 32.5 & 17.6 & 18.9 & 22.6 &
SIB \cite{Hu2020Empirical}          & 88.9 & 89.0 & 98.3 & 82.2 & 89.6 \\
DSN* & 28.4 & 21.1 & 25.6 & 29.5 & 25.8 & 24.6 & 25.8 &
OML \cite{li2020online}            & 87.4 & 86.1 & 97.1 & 78.2 & 87.2 \\
ITA* & 31.4 & 23.0 & 28.2 & 35.7 & 27.0 & 28.9 & 29.0 &
RABN \cite{xu2019self}              & 86.8 & 86.5 & 98.0 & 71.5 & 85.7 \\
KD \cite{belal2021knowledge} & 24.6 & 32.2 & \textbf{33.8} & 35.6 & 46.6 & \textbf{57.5} & 46.6 & 
JiGen \cite{carlucci2019domain}     & 84.8 & 81.0 & 97.9 & 79.0 & 85.7 \\
\hline
Sc$\rightarrow$Tg &  & 37.7 &  &  & 57.9 &  & 47.8 & 
Sc$\rightarrow$Tg & 78.4 & 77.9 & 95.3 & 64.5 & 79.0 \\
%
%
SUDA & \textbf{80.1} & \textbf{76.1} & 25.9 & \textbf{96.0} & \textbf{82.8} & 49.8 & \textbf{68.4} & 
SUDA & \textbf{90.8} & \textbf{89.5} & \textbf{98.8} & \textbf{85.2} & \textbf{91.1} \\
C-SUDA & \textbf{79.9} & \textbf{77.1} & 25.1 & \textbf{95.6} & \textbf{83.2} & 47.6 & \textbf{68.1} & 
C-SUDA & \textbf{89.5} & 88.4 & 97.6 & \textbf{84.6} & \textbf{90.0} \\
\hline
\textit{SUDA$^\star$} & \textit{} & \textit{56.5} & \textit{} & \textit{} & \textit{47.8} & \textit{} & \textit{52.2} & 
\textit{SUDA$^\star$} & \textit{69.1} & \textit{62.2} & \textit{34.1} & \textit{35.5} & \textit{50.2} \\
\textit{C-SUDA$^\star$} & \textit{} & \textit{\textbf{91.2}} & \textit{} & \textit{} & \textit{\textbf{74.1}} & \textit{} & \textit{\textbf{82.7}} & 
\textit{C-SUDA$^\star$} & \textit{\textbf{81.9}} & \textit{\textbf{78.5}} & \textit{\textbf{68.9}} & \textit{\textbf{66.1}} & \textit{\textbf{73.8}} \\
\hline
\end{tabular}
\caption{Classification accuracy on PACS with ResNet18.
* results are taken from \cite{gholami2020unsupervised}.
Legend:  \textit{
\textbf{Sc$\rightarrow$Tg}: Inferred pseudo-labels, 
\textbf{SUDA:} Source-free UDA, \textbf{C-SUDA:} Continual source-free UDA, \textbf{$\star$:} Accuracy on real-source, 
\textcolor{blue}{\textbf{A}}: Art-Painting, \textcolor{blue}{\textbf{C}}: Cartoon, \textcolor{blue}{\textbf{P}}: Photo, and \textcolor{blue}{\textbf{S}}: Sketch.} 
}
\vspace{-1.5em}
\label{tab:PACS}
\end{table*}

%% file: tables/table_domainnet.tex
\setlength{\tabcolsep}{10.9pt}
\begin{table}[!t]
\centering
\scriptsize
\begin{tabular}{|l|c c c c c c |c|}
\hline
\multicolumn{8}{|c|}{\textbf{Multi-Source C-SUDA}} \\
\hline
Tg & 
\textcolor{blue}{\textbf{\textit{C}}} &
\textcolor{blue}{\textbf{\textit{I}}} &
\textcolor{blue}{\textbf{\textit{P}}} &
\textcolor{blue}{\textbf{\textit{Q}}} &
\textcolor{blue}{\textbf{\textit{R}}} &
\textcolor{blue}{\textbf{\textit{S}}} &
Avg.
\\
\hline
MM \cite{peng2019moment} & 58.6 & 26.0 & 52.3 & 6.3 & 62.7 & 49.5 & 42.6 \\
OML \cite{li2020online}  & 62.8 & 21.3 & 50.5 & 15.4 & 64.5 & 50.4 & 44.1 \\
CMSS \cite{yang2020curriculum} & 64.2 & 28.0 & 53.6 & 16.0 & 63.4 & 53.8 & 46.5 \\
DRT+ST \cite{li2021dynamic} & 71.0 & \textbf{31.6} & \textbf{61.0} & 12.3 & \textbf{71.4} & 60.7 & 51.3 \\
\hline
Sc$\rightarrow$Tg & 68.5 & 23.6 & 53.5 & 17.6 & 65.9 & 55.2 & 47.4  \\
%
SUDA & 70.8 & 27.2 & 58.1 & \textbf{24.1} & 69.5 & 60.1 & \textbf{51.6} \\
C-SUDA & \textbf{71.4} & 26.5 & 57.1 & \textbf{24.2} & 67.9 & 59.0 & 51.0 \\
\hline
\textit{SUDA$^\star$} & \textit{29.6} & \textit{30.7} & \textit{35.3} & \textit{7.9} & \textit{35.1} & \textit{36.7} & \textit{29.2} \\
\textit{C-SUDA$^\star$} & \textit{\textbf{64.5}} & \textit{\textbf{67.0}} & \textit{\textbf{63.3}} & \textit{\textbf{55.8}} & \textit{\textbf{59.9}} & \textit{\textbf{65.2}} & \textit{\textbf{62.6}} \\
\hline
\end{tabular}
\caption{Classification accuracy on DomainNet with ResNet101. 
Legend: \textit{
\textcolor{blue}{\textbf{C}}: Clipart, \textcolor{blue}{\textbf{I}}: Infograph, \textcolor{blue}{\textbf{P}}: Painting, \textcolor{blue}{\textbf{Q}}: Quickdraw, \textcolor{blue}{\textbf{R}}: Real, and \textcolor{blue}{\textbf{S}}: Sketch.}
}
\vspace{-1.5em}
\label{tab:domainnet}
\end{table}

%% file: tables/table_VisdaC.tex
\setlength{\tabcolsep}{1.5pt}
\renewcommand{\arraystretch}{1.2} 

\begin{table*}[!t]
\centering
\scriptsize
\begin{tabular}{|l|c c c c c c c c c c c c |c|}
\hline
Methods & \textcolor{blue}{\textbf{\textit{plane}}} & \textcolor{blue}{\textbf{\textit{bcycl}}} & \textcolor{blue}{\textbf{\textit{bus}}} & \textcolor{blue}{\textbf{\textit{car}}} & \textcolor{blue}{\textbf{\textit{horse}}} & \textcolor{blue}{\textbf{\textit{knife}}} & \textcolor{blue}{\textbf{\textit{mcycl}}} & \textcolor{blue}{\textbf{\textit{person}}} & \textcolor{blue}{\textbf{\textit{plant}}} & \textcolor{blue}{\textbf{\textit{skate}}} & \textcolor{blue}{\textbf{\textit{train}}} & \textcolor{blue}{\textbf{\textit{truck}}} & Avg.  \\
\hline
Inferred & 64.2 & 6.3 & 75.2 & 21.7 & 55.9 & 95.7 & 22.8 & 1.4 & 79.8 & 0.7 & 82.8 & 19.8 & 46.3  \\
\hline
DADA \cite{tang2020discriminative} & 92.9 & 74.2 & 82.5 & 65.0 & 90.9 & \textbf{93.8} & 87.2 & 74.2 & 89.9 & 71.5 & 86.5 & 48.7 & 79.8 \\
SHOT \cite{pmlr-v119-liang20a} & 94.3 & \textbf{88.5} & 80.1 & 57.3 & 93.1 & 94.9 & 80.7 & 80.3 & 91.5 & 89.1 & 86.3 & 58.2 & 82.9 \\
A$^2$Net \cite{Xia_2021_ICCV} & 94.0 & 87.8 & 85.6 & 66.8 & 93.7 & 95.1 & 85.8 & 81.2 & 91.6 & 88.2 & 86.5 & 56.0 & 84.3 \\
\hline
Sc$\rightarrow$Tg & 64.2 & 6.3 & 75.2 & 21.7 & 55.9 & 95.7 & 22.8 & 1.4 & 79.8 & 0.7 & 82.8 & 19.8 & 46.3  \\
%
%
SUDA & \textbf{94.8} & 68.1 & \textbf{89.5} & \textbf{88.1} & 86.5 & 90.4 & \textbf{87.4} & \textbf{89.0} & 53.2 & 81.5 & \textbf{96.9} & \textbf{93.0} & \textbf{84.8}  \\
C-SUDA & \textbf{94.9} & 67.3 & \textbf{89.2} & \textbf{87.8} & 86.1 & 90.0 & 86.6 & \textbf{88.7} & 53.1 & 80.9 & \textbf{96.5} & \textbf{94.6} & \textbf{84.6} \\
\hline
\textit{SUDA$^\star$} & \textit{45.2} & \textit{18.5} & \textit{55.9} & \textit{52.7} & \textit{54.8} & \textit{44.3} & \textit{12.5} & \textit{41.4} & \textit{24.6} & \textit{35.1} & \textit{40.2} & \textit{51.2} & \textit{39.7} \\
\textit{C-SUDA$^\star$} & \textit{\textbf{47.6}} & \textit{\textbf{21.4}} & \textit{\textbf{58.2}} & \textit{\textbf{54.3}} & \textit{\textbf{61.1}} & \textit{\textbf{49.5}} & \textit{\textbf{27.9}} & \textit{\textbf{41.9}} & \textit{\textbf{44.8}} & \textit{\textbf{36.2}} & \textit{\textbf{43.1}} & \textit{\textbf{55.4}} & \textit{\textbf{45.1}} \\
\hline
\end{tabular}
\caption{Classification accuracy on Visda-C with ResNet101.
}
\vspace{-2.5em}
\label{tab:VisdaC}
\end{table*}